%% file: main.tex
\documentclass[conference,a4paper]{IEEEtran}
\usepackage[T1]{fontenc}
\usepackage[utf8]{inputenc}
\usepackage{amsmath,booktabs,tabularx,url,graphicx,cite}
\usepackage[hidelinks]{hyperref}
\begin{document}
\title{CoToan: Executable Rules and a Representation Contract for Reproducible Game AI Evaluation}
\IfFileExists{authors.tex}{\input{authors.tex}}{\author{}}
\maketitle
\begin{abstract}
Game AI results are comparable only when different implementations expose the same decision problem. Yet seemingly minor representation choices can silently change that problem by altering action probabilities or collapsing states with different outcomes. We study this issue through CoToan, an executable specification and representation contract for Vietnamese Mathematical Chess connecting the Game Description Language (GDL) with OpenSpiel. The game provides a useful case because a single capture may have multiple arithmetic justifications, while outcomes can depend on scoring rules and a finite playing horizon. CoToan therefore represents actions by their state transitions rather than by arithmetic witnesses, and treats time and victory configuration as part of task identity. Cross-representation tests show agreement in legal actions, successor states, and terminal outcomes across hundreds of sampled and constructed positions, complemented by exhaustive checks of all 32 operator masks and 512 opposing-material subsets. Controlled diagnostics further show that uniform witness-level sampling alters endpoint-action probabilities, while omitting remaining time or victory settings can merge states with different terminal behavior. These results show that action semantics, time, and victory configuration are not merely implementation details. They are part of the experimental task itself and must be represented explicitly for reproducible game AI.

\end{abstract}
\section{Introduction}
\label{sec:1}

Vietnamese Mathematical Chess is a Vietnamese arithmetic board game with distinctive movement and capture rules \cite{vnu2015}. We use it as a case study for a broader reproducibility problem in game AI: when do different implementations actually define the same decision problem?

The game exposes two representation issues that generalize beyond its particular rules. First, a single capture can have multiple arithmetic justifications even though all of them produce the same state transition. If these justifications are represented as separate actions, an implementation can change the probability of selecting a move without changing the set of reachable board positions. Second, identical board positions can have different terminal behavior under different time or victory settings. An environment that omits these variables can therefore merge states that require different outcomes. These examples illustrate a broader point: implementation choices that appear representational can silently change the task presented to an agent.

A representation contract must therefore specify not only which moves and transitions are legal, but also which distinctions are part of the agent's decision problem. This matters when comparing search and learning systems. If two environments differ in action semantics, time information or victory configuration, differences in agent behavior cannot be attributed cleanly to the algorithm alone.

CoToan\footnote{\url{https://github.com/nguyenthanhasia/cotoan}} addresses this problem through versioned executable rules, a declarative Game Description Language (GDL) description, an OpenSpiel environment and reproducible conformance diagnostics. The design separates three concerns: the mechanics and settings that define a task, the arithmetic witnesses that explain why an action is legal, and the interface through which an agent observes and acts on that task. Our contributions are:

C1. Executable task definition. We specify arithmetic capture, path blocking, capture-value scoring and ordered terminal conditions as an executable and configurable task definition. The specification explicitly separates invariant movement mechanics from victory settings that alter task semantics.

C2. Cross-representation contract. We define a correspondence between declarative GDL states and actions and the sequential OpenSpiel interface, preserving legal actions, state transitions, time and configuration identity across representations. Arithmetic witnesses are retained as explanatory metadata rather than exposed as additional agent actions.

C3. Conformance evidence and representation diagnostics. We compare a separately written move predicate, a declarative GDL realization and the OpenSpiel interface across sampled, constructed and configuration-varying states. Controlled diagnostics isolate probability distortion under uniform witness-level sampling and state aliasing caused by omitted time or victory settings.

We investigate three research questions. RQ1 asks whether the GDL, executable-rule and OpenSpiel representations agree on legal actions, successor states and outcomes across supported configurations. RQ2 asks whether representing arithmetic witnesses as distinct actions changes action sampling even when the resulting state transitions are identical. RQ3 asks whether board occupancy and the player to move are sufficient state variables under finite-horizon and configurable victory rules.

Together, these questions identify three requirements for decision-problem equivalence: consistent state transitions, unambiguous action identity and sufficient state information for adjudication. Our focus is reproducibility of the decision problem presented to agents, rather than reproducibility of training outcomes or playing strength.

\section{The Game: Board, Movement and Arithmetic Capture}
\label{sec:2}
\subsection{Board and objective}
\label{sec:2.1}
\begin{figure}[htbp]
\centering
\includegraphics[width=.9\columnwidth]{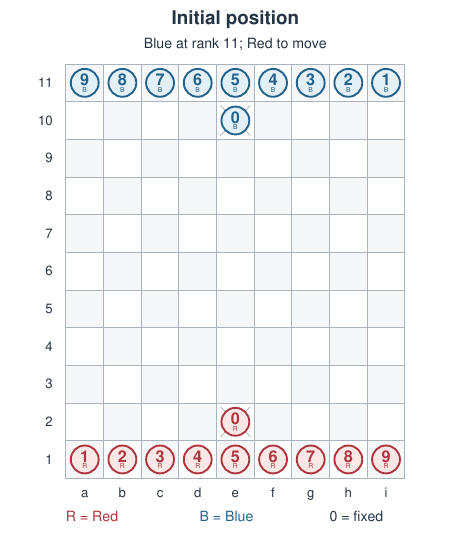}
\caption{Initial arrangement in the research profile. Numerals are piece values; R and B identify the two sides independently of color. The crossed squares e2 and e10 are the fixed zero locations.}
\label{fig:initial-position}
\end{figure}

The board contains nine files and eleven ranks. Each side initially has one piece of each value from 0 through 9, with its zero fixed \cite{vnu2015}. Fig.~\ref{fig:initial-position} fixes the coordinates: Red starts at the lower edge and moves first by default. The players alternate one move at a time.

Capturing the opponent's zero wins. A zero can support a capture but cannot move. Other captures add the target's value to the score and may trigger a configured victory. There is no check, promotion or compulsory capture. The research profile exposes five selectable arithmetic result types: addition, subtraction, multiplication, integer quotient and remainder. All five are enabled by default.

\subsection{Ordinary movement}
\label{sec:2.2}
A nonzero piece of value a can move between one and a squares in any horizontal, vertical or diagonal direction. Every traversed square, including the destination, must be empty. It cannot turn, jump over another piece or capture an enemy by ordinary sliding movement.

\begin{figure}[htbp]
\centering
\includegraphics[width=\columnwidth]{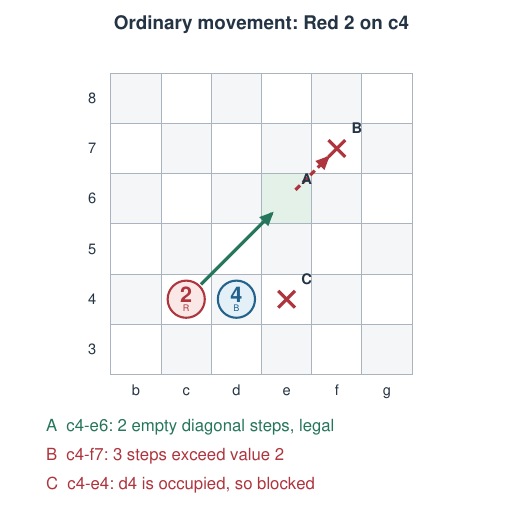}
\caption{Ordinary movement in a local board window. A is legal; B exceeds the piece's range; C is blocked by the enemy on d4. The dashed extension is forbidden. Remote pieces are omitted.}
\label{fig:quiet-movement}
\end{figure}
In Fig.~\ref{fig:quiet-movement}, the red 2 can move from c4 to e6 because the two-step diagonal is clear. The destination f7 exceeds its range. The destination e4 is within range but inaccessible because d4 is occupied. These are distinct geometric reasons for rejecting a move.

\subsection{Arithmetic capture}
\label{sec:2.3}
A capture requires an attacker, an adjacent friendly support and an enemy target on the same straight line. The attacker and support provide the arithmetic operands; the target's value is not an operand. The result of an enabled arithmetic operation determines the target's distance from the support, measured in squares along that line. The result must be positive, and all squares between the support and target must be empty. The attacker replaces the target and the support remains stationary.

\begin{figure}[htbp]
\centering
\includegraphics[width=.8\columnwidth]{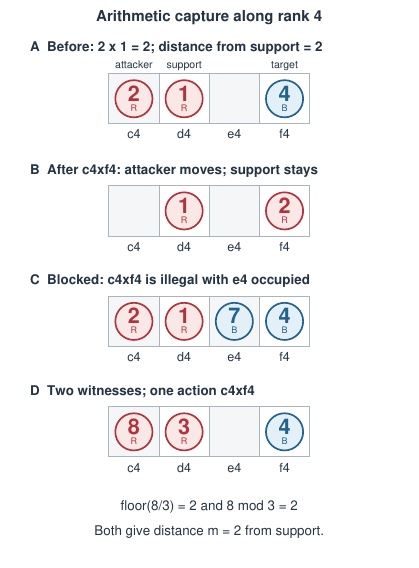}
\caption{Arithmetic capture on rank 4. A and B show a legal move before and after execution. C blocks the target on f4. D gives two arithmetic witnesses for one action. These are constructed local examples; reachability is not assumed.}
\label{fig:arithmetic-capture}
\end{figure}
In Fig.~\ref{fig:arithmetic-capture}A, red 2 on c4 uses red 1 on d4: multiplication gives $2\times1=2$, allowing the attacker to capture the target on f4, two squares beyond the support. Fig.~\ref{fig:arithmetic-capture}B shows the resulting board: c4 is empty, d4 retains red 1 and f4 contains red 2. The blue target is removed and the turn changes.

Fig.~\ref{fig:arithmetic-capture}C inserts blue 7 on e4, blocking the capture of f4. The blocker may nevertheless be captured: subtraction gives 2-1=1, reaching e4 one square beyond the support. Fig.~\ref{fig:arithmetic-capture}D illustrates another issue. Both floor(8/3)=2 and 8 mod 3=2 justify c4xf4. They explain the same action and produce the same successor. Section~\ref{sec:4} makes this distinction part of the action contract, and Section~\ref{sec:5} measures its effect on sampling.

\subsection{Scoring and victory settings}
\label{sec:2.4}
The zero is captured by the same procedure: red 2 on e7 with red 1 on e8 can capture blue zero on e10 through multiplication if e9 is empty. This wins regardless of the score.

Scores start at zero and accumulate captured values. A positive threshold enables immediate victory when a player's score reaches or exceeds it; zero disables that condition. At the default research horizon of 300 plies, score adjudication awards victory to the higher score and a draw to equal scores. One ply is one player's move. Threshold victory and horizon adjudication are separate settings; enabling a threshold does not remove the finite research limit.

We evaluate terminal conditions in this order: zero capture, threshold victory, horizon adjudication, then no-move loss. The last condition is an explicit environment convention for an otherwise nonterminal player with no action. An unconditional horizon draw is retained as an optional experimental setting, not the default rule. Clocks and an unbounded-duration mode are not modeled. Every configuration has its own task identifier.

\section{Related Work}
\label{sec:3}
\subsection{Game description and interoperability}
\label{sec:3.1}
PGN and its associated FEN specification support chess game and position interchange \cite{pgn1994}. Such records do not define the arithmetic transition rules of a different game. General Game Playing instead communicates rules independently of the player \cite{ggp2005}. Original GDL describes deterministic, complete-information games; Thielscher's extension addresses randomness and incomplete information \cite{gdl2010}. Our profile requires only the original language, here called GDL-I.

Regular Boardgames offers a regular-language formalism for board games \cite{rbg2019}, while Ludii organizes descriptions through compositional ludemes \cite{ludii2020}. OpenSpiel supplies algorithm-facing game interfaces and research infrastructure \cite{openspiel2019}. These systems serve complementary purposes. We choose GDL for explicit relational semantics and OpenSpiel for integration with search and learning code. CoToan builds on this infrastructure with a game-specific profile and a checked correspondence across the declarative and agent-facing interfaces.

\subsection{Search, learning and reproducible evaluation}
\label{sec:3.2}
Shannon's formulation of computer chess \cite{shannon1950} and Knuth and Moore's analysis of alpha-beta pruning \cite{knuth1975} establish a classical search perspective. Deep Blue demonstrates a highly engineered chess system \cite{deepblue2002}, whereas solving checkers establishes a game-theoretic outcome \cite{checkers2007}. 

UCT provides a bandit-based planning method \cite{uct2006}, and the MCTS literature describes how tree and rollout policies influence behavior \cite{mcts2012}. AlphaGo combined learned policies and values with search \cite{alphago2016}; AlphaGo Zero trained through self-play without human game data \cite{alphagozero2017}; AlphaZero extended a general self-play approach to chess, shogi and Go \cite{alphazero2018}. MuZero instead plans with a learned model \cite{muzero2020}. Together, these approaches motivate an algorithm-independent environment with well-defined legal actions, observations and rewards.

Pardo et al. distinguish task-defining time limits from training cutoffs and explain why omitting remaining time can violate the Markov property \cite{pardo2018}. Our horizon diagnostic instantiates this established issue in a reachable Co toan position. It is not a new general result about finite-horizon learning.

Henderson et al. identify reproducibility and variability problems in deep reinforcement learning \cite{henderson2018}, and Agarwal et al. emphasize uncertainty-aware evaluation \cite{agarwal2021}. Our work addresses an earlier dependency: agreement on the task itself. Inspired by property-based testing \cite{quickcheck2000}, we combine examples with invariants and separately implemented checks. The contribution is a versioned Co toan specification and conformance evidence, not a new general game language or search algorithm.

\section{Executable Specification and Representation Contract}
\label{sec:4}
\subsection{State and legal actions}
\label{sec:4.1}
Let $S=(B,p,n;\theta)$, where $B$ records occupancy, $p$ is the controller and $n$ is elapsed plies. Immutable configuration $\theta=(H,T,h,O,f_0)$ specifies the horizon, score threshold, horizon rule, operator mask and starting player. A well-formed state has unique piece identities per side, anchored surviving zeros, at least one surviving zero and $0\leq n\leq H$. With both zeros present and a positive threshold, both players cannot already have met it: play stops at the first crossing, so such snapshots are rejected. For zero-based file $f$ and rank $r$, the square index is $q=9(r-1)+f$.

Using two-dimensional square coordinates, let a nonzero attacker of value $a$ occupy $s$, and let $d$ be one of eight unit directions. An ordinary move to $s+Ld$ requires $1\leq L\leq a$ and an empty path including the destination. A capture requires friendly support of value $b$ at $s+d$. Let $m$ be the arithmetic result, measured as the distance from the support to the target. A valid capture requires $m\geq1$, an enemy target at $s+(m+1)d$, and empty squares $s+kd$ for $2\leq k\leq m$. The permitted arithmetic results are:

\begin{table}[ht]
\centering\footnotesize
\caption{Arithmetic capture results measured from the support.}
\begin{tabularx}{\columnwidth}{@{}XXX@{}}
\toprule
Operation & Distance $m$ from support & Guard \\
\midrule
Addition & $(a+b)\bmod 10$ & Always \\
Subtraction & $a-b$ & $a>b$ \\
Multiplication & $(ab)\bmod 10$ & Always \\
Division & $\lfloor a/b\rfloor$ & $a>b>0$ \\
Remainder & $a\bmod b$ & $a>b>0$ \\
\bottomrule
\end{tabularx}
\end{table}
All capture results must satisfy $m\geq1$; a zero result would point to the support square itself. The additional step in the target coordinate accounts for the support's position adjacent to the attacker. Division is floored, not restricted to exact divisibility. Only enabled operators contribute witnesses. These are the normative movement rules of this specification; the historical citations provide attribution, not the operational equations. A transition clears $s$, replaces the destination, preserves other cells, toggles $p$ and increments $n$. Ordered adjudication follows Section~\ref{sec:2.4}.

Let $V_j(B)$ contain player $j$'s surviving values. Player $i$'s captured-value score is

\[
C_i(B)=45-\sum_{v\in V_{1-i}(B)}v.
\]
The initial sum is 45; captures only remove pieces and no rule creates them. Scores therefore follow from occupancy without independent history-dependent counters. Settings determine when these points decide the winner.

\subsection{Actions are transitions, not arithmetic witnesses}
\label{sec:4.2}
We identify an action by its source and destination. The GDL term is (move sSRC sDST); the OpenSpiel ID is 99*SRC+DST. The ambient policy space has 9801 IDs, with illegal entries masked. Arithmetic witnesses remain available as explanatory metadata.

This choice takes the quotient of operator-labeled captures by endpoint equality. Within a fixed state, witnesses with the same endpoints share the same attacker, target, support and board update, so collapsing them preserves the successor relation. It does not preserve a policy that samples witnesses uniformly. If an endpoint action a has k(a) witnesses, with k(a)=1 for an ordinary move, the two distributions are:

\[
p_{\mathrm{move}}(a)=\frac{1}{|A(S)|},\qquad p_{\mathrm{witness}}(a)=\frac{k(a)}{\sum_{b\in A(S)}k(b)}.
\]
Thus arithmetic multiplicity is a potential policy confound, even when the displayed board transitions remain legal. The endpoint contract prevents an incidental choice of explanation from becoming an extra decision for the agent.

\subsection{Declarative encoding and one-step correspondence}
\label{sec:4.3}
GDL represents dynamic information with cell, control and ply facts. Static ray, between and first relations encode geometry. Facts tabulate all 90 arithmetic pairs and enabled operators. A finite fold over values 1 through 9 sums missing opposing material using addition-table facts, deriving each score without extra dynamic state. Threshold and score-comparison facts implement the chosen victory settings. No engine-specific arithmetic extension is needed. Candidate actions precede terminal status, avoiding a legal/terminal cycle; domains are finite with no recursion through negation.

GDL uses joint actions, whereas our OpenSpiel environment is sequential. We assign the off-turn role the sole action noop. Let $\Phi$ map a well-formed Python state to its GDL facts, and let $J_p(a)$ pair the controlling player's action with the other player's noop. Write $T_P$ and $T_G$ for the respective transition functions.

Proposition 1 (one-step correspondence). Under the specified rules, for fixed $\theta$ and a well-formed nonterminal state, controlling-player legal actions correspond bijectively to legal GDL joint actions. For each such action,

\[
\Phi\!\left(T_P(S,a)\right)=T_G\!\left(\Phi(S),J_p(a)\right).
\]
Proof sketch. Rays enumerate the permitted directions and distances. Range, support, enabled-operation and occupancy predicates impose the same movement conditions as the imperative generator. GDL set semantics and endpoint deduplication identify the same actions. Both updates clear the source, replace the destination, preserve other cells, toggle the controller and increment the ply. Induction over the nine score-fold stages yields the same captured-value totals. Both realizations then apply the ordered zero, threshold, horizon and mobility tests. Legality, successor states and terminal outcomes therefore correspond. Utility conversion is $u=2g/100-1$ for GDL goals $g\in\{0,50,100\}$.

The argument concerns the representation definitions. The tests below check their executable realizations, including the generated facts and adapter behavior.

\subsection{Observation and framework contract}
\label{sec:4.4}
The registered OpenSpiel game, python\_cotoan, accepts all five configuration fields. Player IDs 0 and 1 denote Red and Blue; terminal returns are -1, 0 or 1. The observation contains 20 identity planes, a controller plane, a remaining-budget plane and five configuration scalars, totaling 2183 float32 entries. Colors are absolute; supported horizons 1 through 10000 and the other integer settings are exactly representable.

Time awareness is required by a task-defining horizon \cite{pardo2018}; $H-n$ is one sufficient encoding. For a fixed game, configuration can be implicit in the environment. We additionally expose it so pooled agents and datasets can distinguish tasks. Identity planes, controller, budget and configuration recover the specified state and scores. Version-2 snapshots include the complete configuration and reject mismatched imports. OpenSpiel's game-and-state serialization also preserves history. Structural validation is not a reachability proof.

\subsection{Reusing search agents}
\label{sec:4.5}
An agent need not interpret GDL at every search node. It can use OpenSpiel's legal\_actions, child or clone/apply\_action, is\_terminal and returns methods. Heuristics remain agent components; a separate optimized engine can implement the same contract. Shared state-cache keys must distinguish configuration, board, controller and remaining budget. A port preserves behavior only when rules, evaluation conventions and action handling agree.

\section{Evaluation}
\label{sec:5}
\subsection{Protocol and reproducibility}
\label{sec:5.1}
The conformance experiment uses a ray-based implementation and a separately written exhaustive source/destination predicate. Seed 20260901 produces ten trajectories of up to 30 plies, contributing 300 reachable state samples and 300 transitions. A further 300 structurally valid synthetic positions vary occupancy and piece placement. Samples need not be distinct. Complete move sets, rotation/color-swap behavior and snapshot round trips are compared on all 600 positions.

A second path translates the emitted stratified GDL subset to clingo. At H=12, 70 selected fixtures compare legality and outcomes. One legal action is sampled for each of 66 nonterminal fixtures, and its complete successor fact set is compared. Initial facts and all 90 arithmetic pairs are checked separately. 
The committed H=300 description is regenerated and its initial state and one successor are checked.

The default configuration uses all operators, no score threshold and score adjudication at H=300. OpenSpiel runs 20 simulation tests at H=12. A separate audit varies threshold, horizon rule, operator mask and starter across nine configurations; it also checks all 32 masks and 512 subsets of Blue's nonzero material. Additional round trips check histories, observations and configuration mismatch rejection. An eight-simulation MCTS decision and a depth-two alpha-beta decision check usability, not strength. Figures are generated and checked by the engine. The environment is Python 3.12.13, OpenSpiel 2.0.2 and clingo 5.8.2.


\subsection{RQ1: Cross-representation conformance}
\label{sec:5.2}
\begin{table}[ht]
\centering\footnotesize
\caption{Cross-representation conformance results.}
\begin{tabularx}{\columnwidth}{@{}XXX@{}}
\toprule
Check & Scope & Result \\
\midrule
Unit tests & 41 tests & Passed \\
Independent move predicate & 600 positions; 59707 legal-action instances & No mismatch \\
Symmetry and snapshots & 600 positions & No mismatch \\
GDL legality and outcomes & 70 positions & No mismatch \\
GDL successor facts & 66 transitions & No mismatch \\
Arithmetic facts & 90 value pairs & No mismatch \\
OpenSpiel simulation tests & 20 games, H=12 & Passed \\
Configuration audit & 210 states; 158 successors & No mismatch \\
Additional simulation tests & 45 games; 9 settings & Passed \\
Operator-mask comparisons & 32 masks & No mismatch \\
Relational score fold & 512 material subsets & No mismatch \\
Opening exact-depth counts & Depths 1 and 2 & 87; 7457 \\
\bottomrule
\end{tabularx}
\end{table}
The nine-configuration audit compares 210 valid states and one selected successor for each of 158 nonterminal cases across Python, GDL and OpenSpiel, plus 45 simulation tests. Six dual-threshold snapshots are rejected before comparison. These checks are distinct from the 70-state baseline. Exact-depth counts remain 87 and 7457 at depths one and two. A terminal position contributes no leaves at positive depth; the counts are regression fingerprints, not game-tree size estimates.

\subsection{RQ2: Arithmetic witnesses change action sampling}
\label{sec:5.3}
We re-examine the 66 nonterminal public fixtures, assigning one entry to each ordinary action and one entry per valid arithmetic witness to each capture. Six fixtures contain duplicated endpoint actions. Across the fixtures, 6023 endpoint-action instances become 6031 witness-labeled instances. This is a diagnostic set with constructed and sampled positions, not a prevalence estimate for normal play.

The controlled position in Fig.~\ref{fig:arithmetic-capture}D makes the effect explicit. Including the remote pieces recorded in its fixture, it has 43 endpoint actions and 44 witness-labeled actions. Uniform endpoint sampling selects c4xf4 with probability 1/43; uniform witness sampling selects it with probability 2/44=1/22. The board update is identical, but its probability is nearly doubled. Hence matching legal successors alone is insufficient to guarantee a consistent action-sampling contract.

\subsection{RQ3: Horizon omission aliases reachable states}
\label{sec:5.4}
Starting from the initial position, the legal four-ply cycle a1-a2, i11-i10, a2-a1, i10-i11 restores the board and side to move. Repeating the cycle under H=300 yields two nonterminal states at plies 3 and 299 with identical occupancy, controller and legal actions. Applying i10-i11 to both produces identical boards and controllers again, but the early successor at ply 4 is nonterminal while the successor at ply 300 is a draw.

No capture occurs in this cycle, so both scores remain zero: the late draw follows from equal-score adjudication, not an unconditional draw rule. This instantiates Pardo et al.'s time-awareness issue \cite{pardo2018}. A board-and-controller encoding merges states with different terminal transitions; the budget channel separates them.

Configuration matters independently of time. In a constructed position containing all initial pieces except Blue's 9, Red leads 9-0. At the same limit, score adjudication returns a Red win while the optional draw setting returns a draw. Below the limit, threshold 9 ends the game whereas threshold 0 does not. These configurations share occupancy, controller and ply but not task semantics. The diagnostics establish information requirements, not a measured learning benefit.

\subsection{Limitations and validity}
\label{sec:5.5}
The specification is normative for the supported configurations, not a survey of every historical or tournament variant. No-move loss and unconditional horizon draw are explicitly identified research conventions. The 2015 sources establish attribution and context rather than the operational equations. GDL checks use our subset translator and clingo, not an independent GGP competition runtime. The correspondence argument is not mechanized, and tests do not exhaust reachable states.

Synthetic fixtures and short-horizon simulations target correctness rather than typical strategic behavior. The sampling and aliasing diagnostics establish concrete representation effects but do not quantify their consequences for learning strength, game balance or educational outcomes. A subsequent agent study should match computational budgets, exchange colors, disclose seeds and report uncertainty \cite{henderson2018,agarwal2021}. A horizon or rule change must identify a different task rather than be pooled silently into the same benchmark.

\section{Conclusion}
\label{sec:6}

CoToan establishes an executable representation contract for Vietnamese Mathematical Chess across GDL and OpenSpiel, with no mismatches found in the evaluated cases. The diagnostics show that matching state transitions alone is insufficient: arithmetic witnesses can alter action-sampling probabilities, while omitted time or victory settings can merge states with different terminal outcomes. Reproducible game AI evaluation therefore requires consistent action identity, sufficient state information and explicit task configuration.

\section*{Acknowledgments}

This work is financially supported by VinUniversity under Grant No CECS\_UNI.2026.DA017.

\bibliographystyle{IEEEtran}
\bibliography{references}
\end{document}

%% file: authors.tex
\author{
\IEEEauthorblockN{Nguyen Ha Thanh}
\IEEEauthorblockA{
College of Engineering and Computer Science\\
VinUniversity\\
Hanoi, Vietnam
}
}

%% file: references.bib
@misc{vnu2015,
  author = {Do, Ngoc Diep},
  title = {{Co toan Viet Nam: San pham tri tue mang dam triet li Viet}},
  year = {2015},
  howpublished = {VNU Media, interview, December 4},
  url = {https://vnu.edu.vn/co-toan-viet-nam-san-pham-tri-tue-mang-dam-triet-li-viet-post18673.html},
}

@manual{pgn1994,
  author = {{Interested readers of the Internet newsgroup rec.games.chess}},
  title = {{Portable Game Notation Specification and Implementation Guide}},
  year = {1994},
  note = {Coordinated by Steven J. Edwards},
  url = {https://www.saremba.de/chessgml/standards/pgn/pgn-complete.htm},
}

@article{ggp2005,
  author = {Genesereth, Michael and Love, Nathaniel and Pell, Barney},
  title = {{General Game Playing: Overview of the AAAI Competition}},
  year = {2005},
  journal = {AI Magazine},
  volume = {26},
  number = {2},
  pages = {62--72},
  doi = {10.1609/aimag.v26i2.1813},
  url = {https://doi.org/10.1609/aimag.v26i2.1813},
}

@inproceedings{gdl2010,
  author = {Thielscher, Michael},
  title = {{A General Game Description Language for Incomplete Information Games}},
  year = {2010},
  booktitle = {Proceedings of the AAAI Conference on Artificial Intelligence},
  volume = {24},
  number = {1},
  pages = {994--999},
  doi = {10.1609/aaai.v24i1.7647},
  url = {https://ojs.aaai.org/index.php/AAAI/article/view/7647},
}

@inproceedings{rbg2019,
  author = {Kowalski, Jakub and Mika, Maksymilian and Sutowicz, Jakub and Szykuła, Marek},
  title = {{Regular Boardgames}},
  year = {2019},
  booktitle = {Proceedings of the AAAI Conference on Artificial Intelligence},
  volume = {33},
  number = {1},
  pages = {1699--1706},
  doi = {10.1609/aaai.v33i01.33011699},
  url = {https://ojs.aaai.org/index.php/AAAI/article/view/3991},
}

@inproceedings{ludii2020,
  author = {Piette, Eric and Soemers, Dennis J. N. J. and Stephenson, Matthew and Sironi, Chiara F. and Winands, Mark H. M. and Browne, Cameron},
  title = {{Ludii - The Ludemic General Game System}},
  year = {2020},
  booktitle = {ECAI 2020, Frontiers in Artificial Intelligence and Applications},
  volume = {325},
  pages = {411--418},
  doi = {10.3233/FAIA200120},
  url = {https://doi.org/10.3233/FAIA200120},
}

@misc{openspiel2019,
  author = {Lanctot, Marc and others},
  title = {{OpenSpiel: A Framework for Reinforcement Learning in Games}},
  year = {2019},
  howpublished = {arXiv:1908.09453},
  url = {https://arxiv.org/abs/1908.09453},
  note = {Framework report},
}

@article{shannon1950,
  author = {Shannon, Claude E.},
  title = {{Programming a Computer for Playing Chess}},
  year = {1950},
  journal = {The London, Edinburgh, and Dublin Philosophical Magazine and Journal of Science},
  volume = {41},
  number = {314},
  pages = {256--275},
  doi = {10.1080/14786445008521796},
  url = {https://doi.org/10.1080/14786445008521796},
}

@article{knuth1975,
  author = {Knuth, Donald E. and Moore, Ronald W.},
  title = {{An Analysis of Alpha-Beta Pruning}},
  year = {1975},
  journal = {Artificial Intelligence},
  volume = {6},
  number = {4},
  pages = {293--326},
  doi = {10.1016/0004-3702(75)90019-3},
  url = {https://doi.org/10.1016/0004-3702(75)90019-3},
}

@article{deepblue2002,
  author = {Campbell, Murray and Hoane, Jr., A. Joseph and Hsu, Feng-hsiung},
  title = {{Deep Blue}},
  year = {2002},
  journal = {Artificial Intelligence},
  volume = {134},
  number = {1--2},
  pages = {57--83},
  doi = {10.1016/S0004-3702(01)00129-1},
  url = {https://research.ibm.com/publications/deep-blue},
}

@article{checkers2007,
  author = {Schaeffer, Jonathan and Burch, Neil and Bjornsson, Yngvi and Kishimoto, Akihiro and Muller, Martin and Lake, Rob and Lu, Paul and Sutphen, Steve},
  title = {{Checkers Is Solved}},
  year = {2007},
  journal = {Science},
  volume = {317},
  number = {5844},
  pages = {1518--1522},
  doi = {10.1126/science.1144079},
  url = {https://doi.org/10.1126/science.1144079},
}

@inproceedings{uct2006,
  author = {Kocsis, Levente and Szepesvari, Csaba},
  title = {{Bandit Based Monte-Carlo Planning}},
  year = {2006},
  booktitle = {ECML 2006, Lecture Notes in Computer Science},
  volume = {4212},
  pages = {282--293},
  doi = {10.1007/11871842_29},
  url = {https://link.springer.com/chapter/10.1007/11871842_29},
}

@article{mcts2012,
  author = {Browne, Cameron B. and Powley, Edward and Whitehouse, Daniel and Lucas, Simon M. and Cowling, Peter I. and Rohlfshagen, Philipp and Tavener, Stephen and Perez, Diego and Samothrakis, Spyridon and Colton, Simon},
  title = {{A Survey of Monte Carlo Tree Search Methods}},
  year = {2012},
  journal = {IEEE Transactions on Computational Intelligence and AI in Games},
  volume = {4},
  number = {1},
  pages = {1--43},
  doi = {10.1109/TCIAIG.2012.2186810},
  url = {https://doi.org/10.1109/TCIAIG.2012.2186810},
}

@article{alphago2016,
  title={Mastering the game of Go with deep neural networks and tree search},
  author={Silver, David and Huang, Aja and Maddison, Chris J and Guez, Arthur and Sifre, Laurent and Van Den Driessche, George and Schrittwieser, Julian and Antonoglou, Ioannis and Panneershelvam, Veda and Lanctot, Marc and others},
  journal={nature},
  volume={529},
  number={7587},
  pages={484--489},
  year={2016},
  publisher={Nature Publishing Group UK, London}
}

@article{alphagozero2017,
  title={Mastering the game of go without human knowledge},
  author={Silver, David and Schrittwieser, Julian and Simonyan, Karen and Antonoglou, Ioannis and Huang, Aja and Guez, Arthur and Hubert, Thomas and Baker, Lucas and Lai, Matthew and Bolton, Adrian and others},
  journal={nature},
  volume={550},
  number={7676},
  pages={354--359},
  year={2017},
  publisher={Nature Publishing Group UK, London}
}

@article{alphazero2018,
  title={A general reinforcement learning algorithm that masters chess, shogi, and Go through self-play},
  author={Silver, David and Hubert, Thomas and Schrittwieser, Julian and Antonoglou, Ioannis and Lai, Matthew and Guez, Arthur and Lanctot, Marc and Sifre, Laurent and Kumaran, Dharshan and Graepel, Thore and others},
  journal={Science},
  volume={362},
  number={6419},
  pages={1140--1144},
  year={2018},
  publisher={American Association for the Advancement of Science}
}

@article{muzero2020,
  author = {Schrittwieser, Julian and others},
  title = {{Mastering Atari, Go, Chess and Shogi by Planning with a Learned Model}},
  year = {2020},
  journal = {Nature},
  volume = {588},
  pages = {604--609},
  doi = {10.1038/s41586-020-03051-4},
  url = {https://www.nature.com/articles/s41586-020-03051-4},
}

@inproceedings{pardo2018,
  author = {Pardo, Fabio and Tavakoli, Arash and Levdik, Vitaly and Kormushev, Petar},
  title = {{Time Limits in Reinforcement Learning}},
  year = {2018},
  booktitle = {Proceedings of the 35th International Conference on Machine Learning, Proceedings of Machine Learning Research},
  volume = {80},
  pages = {4045--4054},
  url = {https://proceedings.mlr.press/v80/pardo18a.html},
}

@inproceedings{henderson2018,
  author = {Henderson, Peter and Islam, Riashat and Bachman, Philip and Pineau, Joelle and Precup, Doina and Meger, David},
  title = {{Deep Reinforcement Learning That Matters}},
  year = {2018},
  booktitle = {Proceedings of the AAAI Conference on Artificial Intelligence},
  volume = {32},
  number = {1},
  doi = {10.1609/aaai.v32i1.11694},
  url = {https://ojs.aaai.org/index.php/aaai/article/view/11694},
}

@inproceedings{agarwal2021,
  author = {Agarwal, Rishabh and Schwarzer, Max and Castro, Pablo Samuel and Courville, Aaron C. and Bellemare, Marc G.},
  title = {{Deep Reinforcement Learning at the Edge of the Statistical Precipice}},
  year = {2021},
  booktitle = {Advances in Neural Information Processing Systems},
  volume = {34},
  pages = {29304--29320},
  url = {https://proceedings.neurips.cc/paper/2021/hash/f514cec81cb148559cf475e7426eed5e-Abstract.html},
}

@inproceedings{quickcheck2000,
  author = {Claessen, Koen and Hughes, John},
  title = {{QuickCheck: A Lightweight Tool for Random Testing of Haskell Programs}},
  year = {2000},
  booktitle = {Proceedings of the ACM SIGPLAN International Conference on Functional Programming},
  pages = {268--279},
  doi = {10.1145/351240.351266},
  url = {https://doi.org/10.1145/351240.351266},
}
